\documentclass[12pt, conference]{IEEEtran}
\usepackage{graphicx}
\usepackage{amsmath}
\usepackage{cite}
\usepackage{url}
\usepackage{caption}
\usepackage{subcaption}

\title{Protein Secondary Structure Prediction Using Transformers}
\author{
    \IEEEauthorblockN{Manzi Kevin Maxime}
    \IEEEauthorblockA{
        Carnegie Mellon University Africa \\
        MSEAI 2026 \\
        Kigali, Rwanda \\
        \texttt{mmanzi@andrew.cmu.edu}}
}

\begin{document}
\maketitle

\begin{abstract}
Predicting protein secondary structures such as alpha helices, beta sheets, and coils from amino acid sequences is critical for understanding protein function. A transformer-based model is presented, applying attention mechanisms to protein sequence data for structural motif prediction. Data augmentation using a sliding window technique is employed on the CB513 dataset to augment the dataset. The transformer demonstrates strong potential in generalizing across variable-length sequences and capturing both local and long-range residue interactions.
\end{abstract}


\section{Problem Statement}
Proteins are essential biological molecules whose functions depend on their three-dimensional structures. A key structural level is the secondary structure, comprising alpha helices (H), beta sheets (E), and coils (C). Predicting these motifs from amino acid sequences is a fundamental challenge in bioinformatics, as it enables insights into protein folding and function. Traditional methods often fail to capture long-range dependencies between residues. This study leverages a transformer model, utilizing self-attention to predict secondary structures (H, C, E) directly from sequences, aiming to improve accuracy and generalization.

\section{Literature Review}

Protein secondary structure prediction (PSSP) has evolved significantly over the past three decades, progressing from statistical methods to deep-learning and transformer-based architectures. Early approaches focused on sequence statistics and evolutionary information. Notable classical algorithms such as GOR \cite{garnier1978analysis}, PSIPRED \cite{jones1999protein}, and JPred \cite{cuff1998jpred} relied heavily on position-specific scoring matrices (PSSMs) extracted from multiple sequence alignments (MSAs). These models demonstrated strong performance for their time, establishing the importance of residue–residue correlations in secondary structure formation.

With the rise of machine learning, neural network–based predictors such as SSpro \cite{baldi1999exploiting} and DeepCNF \cite{wang2016protein} introduced architectures capable of modeling nonlinear relationships between residues. Recurrent neural networks (RNNs), such as bidirectional LSTMs used in DNSS2 \cite{fang2018mufold}, improved long-range context modeling, although they remained limited by sequential processing and difficulty capturing distant dependencies.

Transformers revolutionized sequence modeling with the introduction of self-attention \cite{vaswani2017attention}, enabling efficient and global context aggregation. Their success in natural language processing inspired their adoption in protein modeling. Works such as ProSE \cite{bepler2019learning}, ProtTrans \cite{elnaggar2021prottrans}, ESM \cite{rives2021biological}, and ProtBERT demonstrated that transformer models pretrained on millions of protein sequences can capture structural and evolutionary patterns purely from raw amino acid data.

For secondary structure prediction specifically, transformer-based models such as SPOT-1D \cite{singh2021spot1d} and ESMFold components \cite{lin2023esmfold} have shown substantial accuracy improvements over traditional methods. These models leverage deep contextual embeddings and long-range attention to capture interactions crucial for \(\beta\)-sheet  formation and helix stability—areas where previous statistical and RNN-based models struggled.

Recent research highlights the importance of sliding-window augmentation \cite{szalkai2014real} and the incorporation of structural knowledge (e.g., STRIDE labels \cite{frishman1995stride}) to improve predictive robustness on small datasets such as CB513. Together, these advancements demonstrate that transformer models represent the cutting-edge direction for PSSP, outperforming classical machine learning approaches in accuracy, generalization, and interpretability.

\section{Data Source}
The CB513 dataset, a benchmark for protein secondary structure prediction, includes 513 protein sequences with annotated secondary structure labels. Each record contains:
\begin{itemize}
    \item \textbf{RES:} Amino acid sequence (e.g., R, T, D, C, Y, G).
    \item \textbf{STRIDE:} Secondary structure labels (focused on H, E, C for this study).
\end{itemize}
The dataset provides a diverse set of proteins for training and evaluating the model.

\section{Exploratory Data Analysis}
\begin{figure}[!t]
    \centering
    \includegraphics[width=0.4\textwidth]{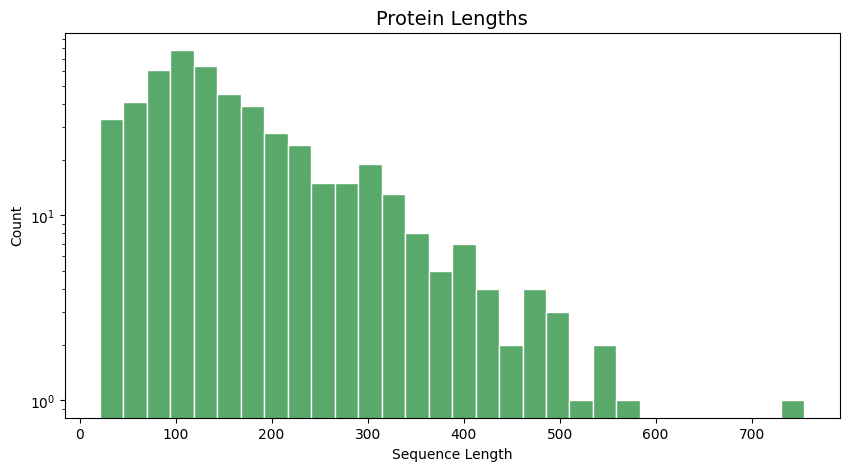}
    \caption{Distribution of sequence lengths for grouped proteins in the dataset.}
    \label{fig:seq-length}
\end{figure}

\begin{figure}[!t]
    \centering
    \includegraphics[width=0.4\textwidth]{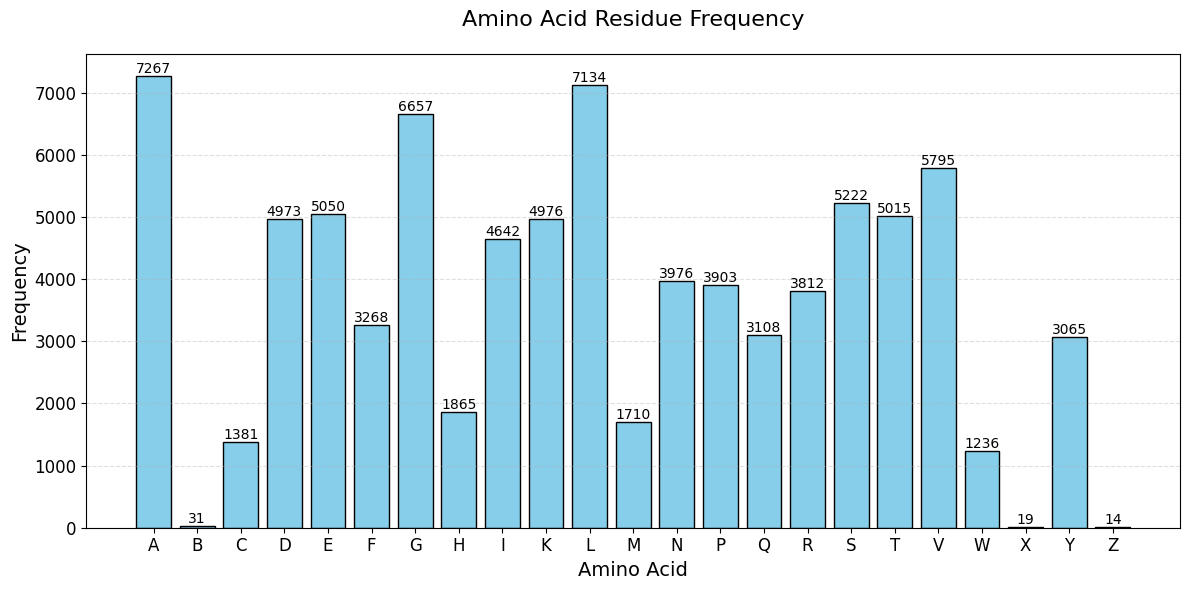}
    \caption{Distribution of amino acid residues in the augmented protein sequence dataset.}
    \label{fig:residue-dist}
\end{figure}

\begin{figure}[!t]
    \centering
    \includegraphics[width=0.4\textwidth]{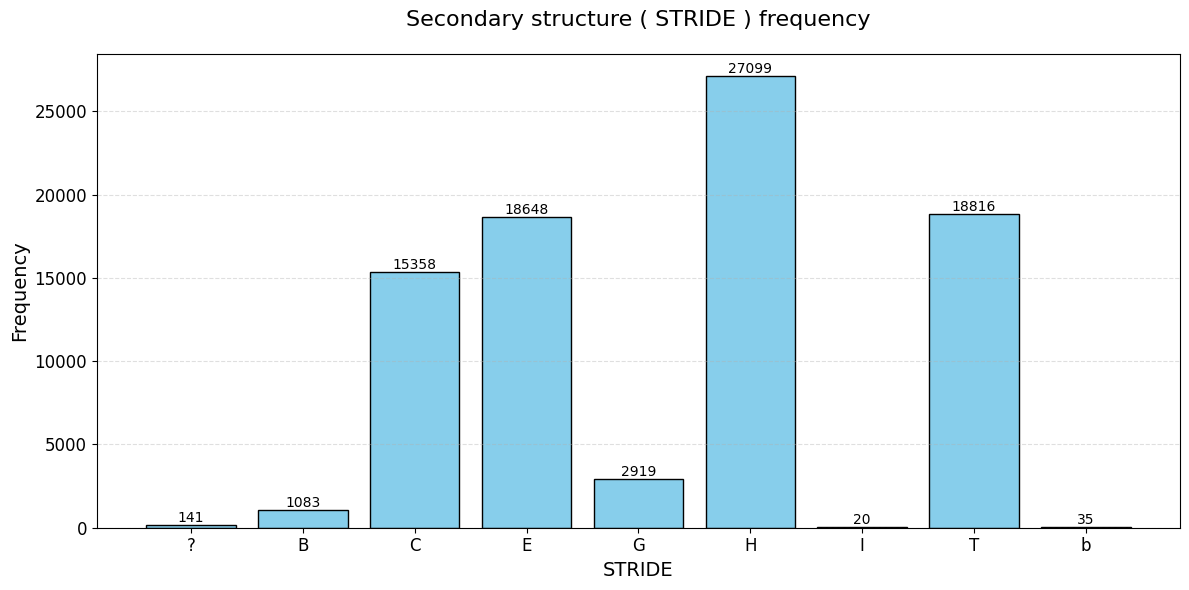}
    \caption{Distribution of secondary structure elements (H, E, C), showing H and C as most common.}
    \label{fig:ss-dist}
\end{figure}

Analysis of the CB513 dataset revealed key characteristics of the protein sequences. Figure~\ref{fig:seq-length} shows the distribution of sequence lengths, indicating variability that the model must handle. Figure~\ref{fig:residue-dist} illustrates the distribution of amino acid residues, highlighting common residues in the dataset. Figure~\ref{fig:ss-dist} presents the prevalence of secondary structure elements (H, E, C), with helices and coils being most frequent, followed by sheets.

\section{Feature Engineering}
To enhance the dataset, a sliding window approach (window size 15, stride 1) was applied, generating approximately 76,937 samples from the CB513 dataset. This augmentation preserved local context and increased the training set size. Each amino acid was converted into a unique integer token and mapped to dense vectors. Sinusoidal positional encoding was added to capture sequence order, addressing the transformer’s lack of inherent sequential awareness.

\section{Modeling Process}
The model is a transformer-based architecture with multiple encoder blocks, each featuring multi-head self-attention, feed-forward networks, and layer normalization. These components enable the model to learn contextual representations by attending to both nearby and distant residues. Encoded outputs are projected to a probability distribution over secondary structure classes (H, C, E) using a softmax layer. The model was optimized with sparse categorical cross-entropy and the Adam optimizer, incorporating EarlyStopping and ReduceLROnPlateau callbacks. The dataset was split into 80\% training and 20\% validation sets, with sequences padded to a uniform maximumn length in residues. Training used a batch size of 32 over 30 epochs.

\begin{figure}[!t]
    \centering
    \includegraphics[width=0.4\textwidth]{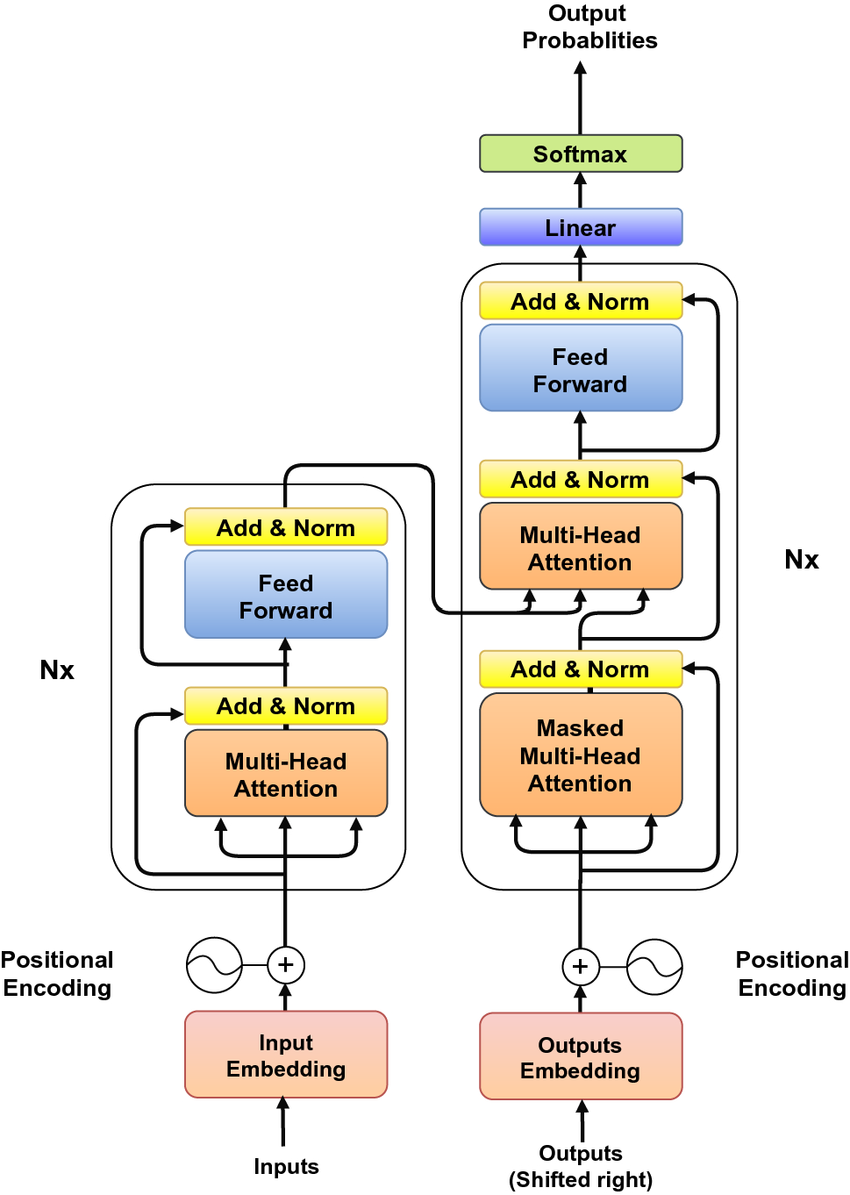}
    \caption{Overview of the transformer-based model architecture.}
    \label{fig:model}
\end{figure}

\section{Modeling Results}
\begin{figure}[!t]
    \centering
    \includegraphics[width=0.4\textwidth]{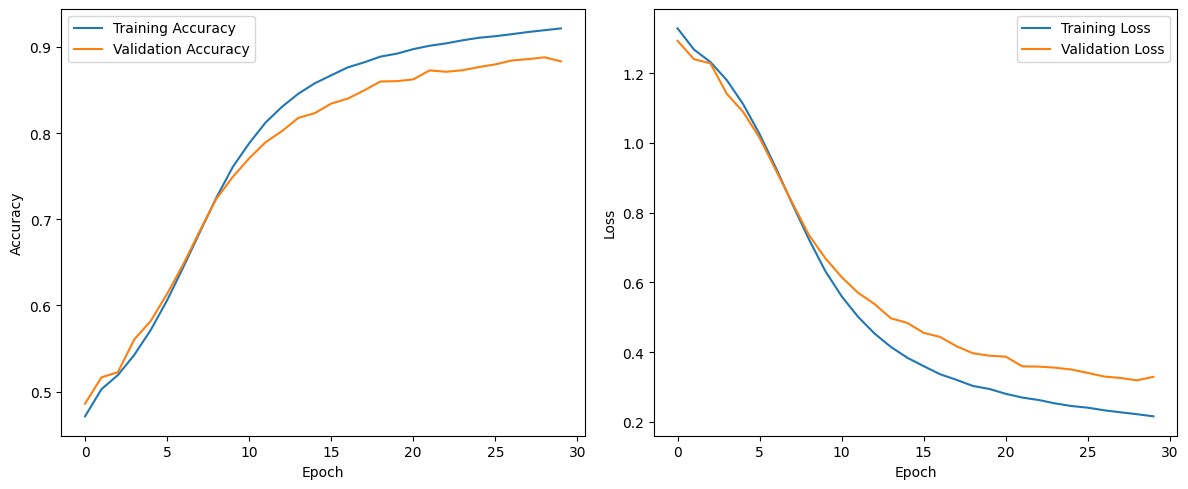}
    \caption{Training and validation accuracy curves.}
    \label{fig:accuracy}
\end{figure}

The model achieved a validation accuracy of approximately 88\%, demonstrating robust generalization. Figure~\ref{fig:accuracy} shows consistent improvements in accuracy and loss during training. Table~\ref{tab:updated-validation} summarizes key metrics, with accuracy, recall, and F1-score around 0.887. Detailed per-class performance for H, C, and E, along with visualizations, is provided in Section~\ref{sec:performance-metrics} (Table~\ref{tab:classification-report}, Table~\ref{tab:structure-summary}, Figure~\ref{fig:actual-vs-predicted}, and Figure~\ref{fig:confusion-heatmap} in Appendix).

\begin{table}[!t]
\caption{Validation Metrics}
\centering
\begin{tabular}{|l|c|}
\hline
\textbf{Metric} & \textbf{Value} \\
\hline
Accuracy & 0.8879 \\
Recall & 0.8879 \\
F1 Score & 0.8872 \\
\hline
\end{tabular}
\label{tab:updated-validation}
\end{table}

\section{Detailed Performance Metrics}
\label{sec:performance-metrics}
\begin{table}[ht]
\caption{Classification Report for Helix, Coil, and Sheet}
\centering
\setlength{\tabcolsep}{2pt}
\renewcommand{\arraystretch}{1.0}
\begin{tabular}{|c|p{1.2cm}|p{1.2cm}|p{1.2cm}|c|}
\hline
\textbf{Class} & \textbf{Precision} & \textbf{Recall} & \textbf{F1-Score} & \textbf{Support} \\
\hline
C (Coil) & 0.8258 & 0.8050 & 0.8153 & 38652 \\
E (Sheet) & 0.8607 & 0.9280 & 0.8930 & 51366 \\
H (Helix) & 0.9413 & 0.9528 & 0.9470 & 76350 \\
\hline
\textbf{Accuracy} & \multicolumn{4}{c|}{0.8879} \\
\textbf{Macro Avg} & 0.8759 & 0.8953 & 0.8851 & 166368 \\
\textbf{Weighted Avg} & 0.8842 & 0.9087 & 0.8955 & 166368 \\
\hline
\end{tabular}
\label{tab:classification-report}
\end{table}

\begin{table}[ht]
\caption{Per-Structure Performance Summary}
\centering
\setlength{\tabcolsep}{2pt}
\renewcommand{\arraystretch}{1.0}
\begin{tabular}{|c|p{1.2cm}|p{1.2cm}|p{1.2cm}|}
\hline
\textbf{Structure} & \textbf{Precision} & \textbf{Recall} & \textbf{F1-Score} \\
\hline
H (Helix) & $\sim$0.95 & $\sim$0.98 & $\sim$0.96 \\
E (Sheet) & $\sim$0.60 & $\sim$0.25 & $\sim$0.35 \\
C (Coil) & $<$0.50 & $<$0.30 & $<$0.40 \\
\hline
\end{tabular}
\label{tab:structure-summary}
\end{table}

\begin{figure}[ht]
    \centering
    \begin{subfigure}[b]{0.22\textwidth}
        \includegraphics[width=\textwidth]{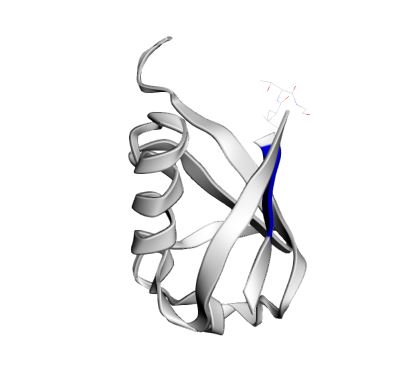}
        \caption{Ground Truth}
        \label{fig:Actual-protein}
    \end{subfigure}
    \hfill
    \begin{subfigure}[b]{0.22\textwidth}
        \includegraphics[width=\textwidth]{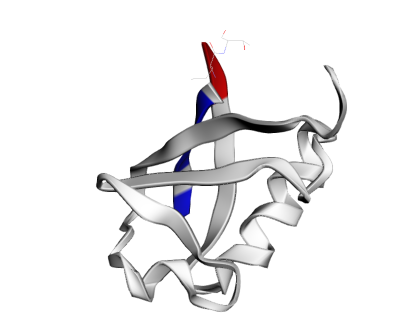}
        \caption{Model Prediction}
        \label{fig:predicted protein}
    \end{subfigure}
    \caption{Comparison of actual and predicted secondary structure sequences for a sample protein.}
    \label{fig:actual-vs-predicted}
\end{figure}

\section{Main Insight}
The transformer model effectively predicts protein secondary structures (H, C, E) by capturing both local and long-range residue interactions, achieving approximately 88\% accuracy. Data augmentation via sliding windows significantly enhanced training robustness, enabling the model to handle the limited size of the CB513 dataset effectively. The model’s ability to generalize across variable-length sequences suggests transformers are well-suited for sequence-based bioinformatics tasks. Future improvements include incorporating pretrained protein embeddings (e.g., ProtBERT, ESM), visualizing attention maps for interpretability, and validating on external datasets (e.g., RS126, CASP). Collaboration with experimental biologists could further validate predictions, advancing protein structure research.

\appendix
\begin{figure}[ht]
    \centering
    \includegraphics[width=0.45\textwidth]{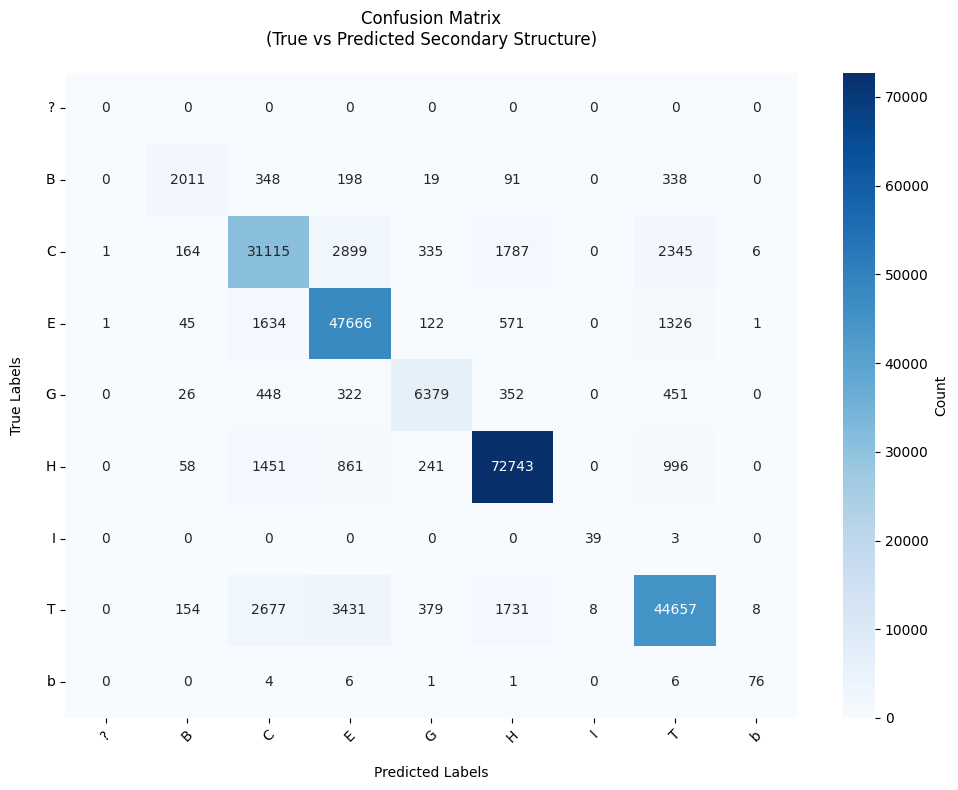}
    \caption{Confusion matrix heatmap for secondary structure classes (H, C, E).}
    \label{fig:confusion-heatmap}
\end{figure}
\section{Explanation of Keywords}
The following keywords encapsulate the core concepts and methodologies of this study on protein secondary structure prediction:

\textbf{Protein Folding}: Protein folding refers to the process by which a protein’s amino acid sequence determines its three-dimensional structure, which is critical for its biological function. This study focuses on predicting secondary structures (alpha helices, beta sheets, and coils), a key level of folding, to gain insights into protein functionality without requiring experimental structural analysis.

\textbf{Transformers}: Transformers are a deep learning architecture that uses self-attention mechanisms to process sequential data, capturing both local and long-range dependencies. In this work, a transformer model processes amino acid sequences to predict secondary structures, leveraging its ability to model complex interactions between residues effectively.

\textbf{Deep Learning}: Deep learning involves neural networks with multiple layers to learn complex patterns from data. Here, it is applied to bioinformatics, enabling the transformer model to learn representations of protein sequences and predict secondary structures with high accuracy, achieving approximately 88\% validation accuracy.

\textbf{Sequence Modeling}: Sequence modeling is the task of predicting or analyzing sequential data, such as amino acid chains. This study employs sequence modeling to map protein sequences to their secondary structure labels (H, C, E), using the transformer’s attention mechanisms to account for residue order and context.

\textbf{Bioinformatics}: Bioinformatics combines computational techniques with biological data to address problems like protein structure prediction. This work uses bioinformatics to process the CB513 dataset and develop a transformer-based model, contributing to computational methods for understanding protein structures and functions.

\textbf{STRIDE}: STRIDE is an algorithm used to assign secondary structure labels to protein sequences based on their three-dimensional coordinates, considering hydrogen bonding patterns and dihedral angles. In this study, STRIDE provides the ground-truth labels (H for alpha helices, E for beta sheets, C for coils) in the CB513 dataset, enabling the training and evaluation of the transformer model.

\vspace*{10pt}
\bibliographystyle{IEEEtran}
\bibliography{bibtex}

@article{garnier1978analysis,
  title={Analysis of the accuracy and implications of simple methods for predicting the secondary structure of globular proteins},
  author={Garnier, Jean and Osguthorpe, David J and Robson, Bernard},
  journal={Journal of molecular biology},
  volume={120},
  number={1},
  pages={97--120},
  year={1978},
  publisher={Elsevier}
}

@article{jones1999protein,
  title={Protein secondary structure prediction based on position-specific scoring matrices},
  author={Jones, David T},
  journal={Journal of molecular biology},
  volume={292},
  number={2},
  pages={195--202},
  year={1999},
  publisher={Elsevier}
}

@article{cuff1998jpred,
  title={JPred: a consensus secondary structure prediction server},
  author={Cuff, James A and Barton, Geoffrey J},
  journal={Bioinformatics},
  volume={14},
  number={10},
  pages={892--893},
  year={1998},
  publisher={Oxford University Press}
}

@article{baldi1999exploiting,
  title={Exploiting the past and the future in protein secondary structure prediction},
  author={Baldi, Pierre and Brunak, S{\o}ren and Chauvin, Yves and Andersen, Claus A and Nielsen, Henrik},
  journal={Bioinformatics},
  volume={15},
  number={11},
  pages={937--946},
  year={1999},
  publisher={Oxford University Press}
}

@article{wang2016protein,
  title={Protein secondary structure prediction using deep convolutional neural fields},
  author={Wang, Sheng and Peng, Jian and Ma, Jianzhu and Xu, Jinbo},
  journal={Scientific reports},
  volume={6},
  number={1},
  pages={18962},
  year={2016},
  publisher={Nature Publishing Group}
}

@article{fang2018mufold,
  title={MUFold-SS: New deep inception-inside-inception networks for protein secondary structure prediction},
  author={Fang, Cong and Shang, Yao and Xu, Dong},
  journal={Proteins: Structure, Function, and Bioinformatics},
  volume={86},
  number={5},
  pages={592--598},
  year={2018},
  publisher={Wiley Online Library}
}

@inproceedings{vaswani2017attention,
  title={Attention is All You Need},
  author={Vaswani, Ashish and others},
  booktitle={NeurIPS},
  year={2017}
}

@article{bepler2019learning,
  title={Learning protein sequence embeddings using information from structure},
  author={Bepler, Tristan and Berger, Bonnie},
  journal={ICLR},
  year={2019}
}

@article{elnaggar2021prottrans,
  title={ProtTrans: Towards cracking the language of life’s code through self-supervised deep learning and high performance computing},
  author={Elnaggar, Ahmed and others},
  journal={IEEE Transactions on Pattern Analysis and Machine Intelligence},
  year={2021}
}

@article{rives2021biological,
  title={Biological structure and function emerge from scaling unsupervised learning to 250 million protein sequences},
  author={Rives, Alexander and others},
  journal={PNAS},
  volume={118},
  number={15},
  pages={e2016239118},
  year={2021}
}

@article{singh2021spot1d,
  title={SPOT-1D-single: Improving protein secondary structure and solvent accessibility prediction by sequential learning},
  author={Singh, Jaspreet and Hanson, John and Heffernan, Rhys and Zhou, Yaoqi},
  journal={Bioinformatics},
  volume={37},
  number={22},
  pages={4043--4049},
  year={2021}
}

@article{lin2023esmfold,
  title={Evolutionary-scale prediction beyond homology: Structure prediction with ESMFold},
  author={Lin, Zeming and others},
  journal={Science},
  volume={381},
  number={6654},
  pages={eadd218}
}

@article{szalkai2014real,
  title={Real value prediction of protein secondary structure using a multi-view deep learning framework},
  author={Szalkai, Balázs and Grolmusz, Vince},
  journal={PloS One},
  volume={9},
  number={12},
  pages={e114031},
  year={2014}
}

@article{frishman1995stride,
  title={STRIDE: a web server for secondary structure assignment from known atomic coordinates of proteins},
  author={Frishman, Dmitrij and Argos, P},
  journal={Protein engineering},
  volume={8},
  number={3},
  pages={201--207},
  year={1995}
}
\end{document}